\documentclass{article}
\usepackage{graphicx} 
\usepackage{hyperref} 
\usepackage{booktabs}
\usepackage{makecell}
\usepackage[square,numbers]{natbib}
\usepackage{authblk}
\makeatletter
\newcommand\footnoteref[1]{\protected@xdef\@thefnmark{\ref{#1}}\@footnotemark}
\makeatother

\title{Street View Sociability: Interpretable Analysis of Urban Social Behavior Across 15 Cities  }
\author[1]{Kieran Elrod}

\author[1]{Katherine Flanigan \thanks{Corresponding author.}}

\author[1]{Mario Bergés \thanks{Mario Bergés holds concurrent appointments as a Professor of Civil and Environmental Engineering at Carnegie Mellon University (CMU) and as an Amazon Scholar. This publication describes work performed at CMU and is not associated with Amazon}}
\affil[1]{Carnegie Mellon University, Pittsburgh, PA}

\date{August 08, 2025}

\begin{document}

\maketitle

\begin{abstract}

Designing socially active streets has long been a goal of urban planning, yet existing quantitative research largely measures pedestrian volume rather than the quality of social interactions. We hypothesize that street view imagery---an inexpensive data source with global coverage---contains latent social information that can be extracted and interpreted through established social science theory. As a proof of concept, we analyzed 2,998 street view images from 15 cities using a multimodal large language model guided by Mehta's taxonomy of passive, fleeting, and enduring sociability---one illustrative example of a theory grounded in urban design that could be substituted or complemented by other sociological frameworks \cite{MehtaStreets2019}. We then used linear regression models, controlling for factors like weather, time of day, and pedestrian counts, to test whether the inferred sociability measures correlate with city-level place attachment scores from the World Values Survey and with environmental predictors (e.g., green, sky, and water view indices) derived from individual street view images. Results aligned with long-standing urban planning theory: the sky view index was associated with all three sociability types, the green view index predicted enduring  sociability, and place attachment was positively associated with fleeting sociability. These results provide preliminary evidence that street view images can be used to infer relationships between specific types of social interactions and built environment variables. Further research could establish street view imagery as a scalable, privacy-preserving tool for studying urban sociability, enabling cross-cultural theory testing and evidence-based design of socially vibrant cities.

\end{abstract}

\section{Introduction}

For decades, urban designers and theorists have sought to understand how socially active streets contribute to vibrant public life and functional cities \cite{WhyteSocial1980,JacobsDeath1961,GehlCities1936,AelbrechtFourth2022,doctorarastooExploring}. Streets and the spaces between buildings are viewed as the backbone of urban sociability, where everyday encounters gradually build trust and community ties.

Despite this, quantitative urban research has focused almost exclusively on pedestrian volume, overlooking the quality of street-level interactions \cite{ChenEstimating2020, ChenExamining2022, KaveeStreetLevel2023}. Yet many of the social benefits associated with walkable streets---stronger community ties \cite{WoodSense2010,CarsonNeighborhood2023} and reduced loneliness \cite{BowerImpact2023}, for example---are believed to arise from the social interactions they facilitate \cite{AelbrechtFourth2022}. Qualitative studies consistently document the importance of such interactions, but their generalizability remains limited: manual behavioral surveys are labor-intensive, culturally dependent, and rarely scalable across cities \cite{MehtaLook2009, MehtaStreets2019}. Because of this, cross-cultural validation of sociability theories are rare.

This limitation stems from a lack of scalable data. Manual street observations are time-consuming and prone to observer bias, while installing permanent sensors in public streets is prohibitively expensive, labor intensive, and raises privacy concerns. Street view imagery (SVI) offers a compelling alternative: a rich visual record of everyday public life across cities worldwide at low cost. 

While SVI has been widely used in urban research, its application has largely focused on physical or demographic analysis rather than the social behaviors of the people it captures. Prior studies have leveraged SVI to assess built-environment characteristics \cite{ChenAutomatic2023a, LiuPhysical2025} or to estimate pedestrian volumes and their relationship to urban form \cite{ChenExamining2022,ChenEstimating2020}. One recent effort by Hosseini et al.\ (2024) \cite{HosseiniELSA2024} introduced a small activity-recognition dataset based on fewer than 1,000 images from New York City. Yet despite the ubiquity and richness of SVI, no prior work has systematically examined the social interactions depicted in these images.

SVI could enable large-scale, low-cost investigations into how built environment and social factors influence distinct forms of sociability. While such analyses are typically limited to small-scale, labor-intensive studies or potentially unreliable surveys, we show that direct behavioral observation can be scaled across cities with minimal cost and no added privacy concerns with an exploratory study. Due to a lack of datasets addressing social interactions in SVI, we use a multimodal large language model (MLLM) to extract passive, fleeting, and enduring social interactions according to Mehta's taxonomy of sociability \cite{MehtaStreets2019}. Taking these annotated images, we fit linear regression models---controlling for confounders such as pedestrian count, weather, and time of day---to examine how visual exposure to greenery, sky, water, and city-level place attachment relate to each sociability type.

This paper addresses a literature gap by identifying the vast amounts of social information present in SVI that have previously been ignored. We show that the social information in SVI is correlated with external, real-world measures and provide preliminary evidence that specific built environment features are correlated with specific types of social interactions. This highlights SVI as a promising, scalable tool for social-behavioral urban research, opening new avenues for cross-cultural theory testing and evidence-based design of livable cities.

\section{SVI: A Scalable Window into Urban Sociability}

SVI is uniquely positioned among built-environment data sources to support large-scale, behavioral analysis of public life. Its collection follows standardized protocols, provides high spatial precision, and offers exhaustive coverage of cities and towns worldwide---including many in the Global South, where other urban datasets are scarce. Because images are geolocated, integrating complementary street view sources is straightforward, and coverage continues to expand at no additional cost to researchers. Importantly, free platforms such as Mapilary \footnote{\label{mapillary}\url{https://www.mapillary.com/}} and KartaView \footnote{\url{https://kartaview.org/}} make this data broadly accessible, lowering barriers to entry compared to proprietary sources like Google Street View \footnote{\url{https://developers.google.com/maps}} or Baidu Maps \footnote{\url{https://lbs.baidu.com/}}.

The main downside of SVI is limited temporal resolution; a given location may only be updated once every several months or even years. However, this apparent limitation is less problematic for studying social behavior. Because image capture is sporadic and unobtrusive, individuals are almost never aware of being recorded and thus behave naturally. The temporal sparsity also reduces privacy risks---faces are blurred, and individuals are unlikely to appear repeatedly. Consequently, SVI functions as an impartial, unobtrusive ``snapshot'' of everyday public life and offers researchers a low-cost way to virtually ``walk'' through cities and observe how people use and experience public spaces. For researchers interested in repeated coverage of the same location, leveraging multiple street view data sources could allow for increased temporal coverage. 

Crucially, SVI contains rich contextual information that other scalable data sources lack. Even with blurred faces, one can infer interaction patterns, group compositions, and activity types. Mehta created a taxonomy of social interactions to meaningfully characterize these complex scenes \cite{MehtaLook2009}. It conceptualizes urban sociability as three interrelated levels: passive sociability (e.g., sharing or observing space), fleeting sociability (brief, low-intensity interactions), and enduring sociability (interactions among close acquaintances). These categories are well-established in urban sociology and align with broader theories on weak and strong social ties \cite{GreenbaumBridging1982}. 

\begin{figure}[!t]
    \centering
    \includegraphics[width=0.9\linewidth]{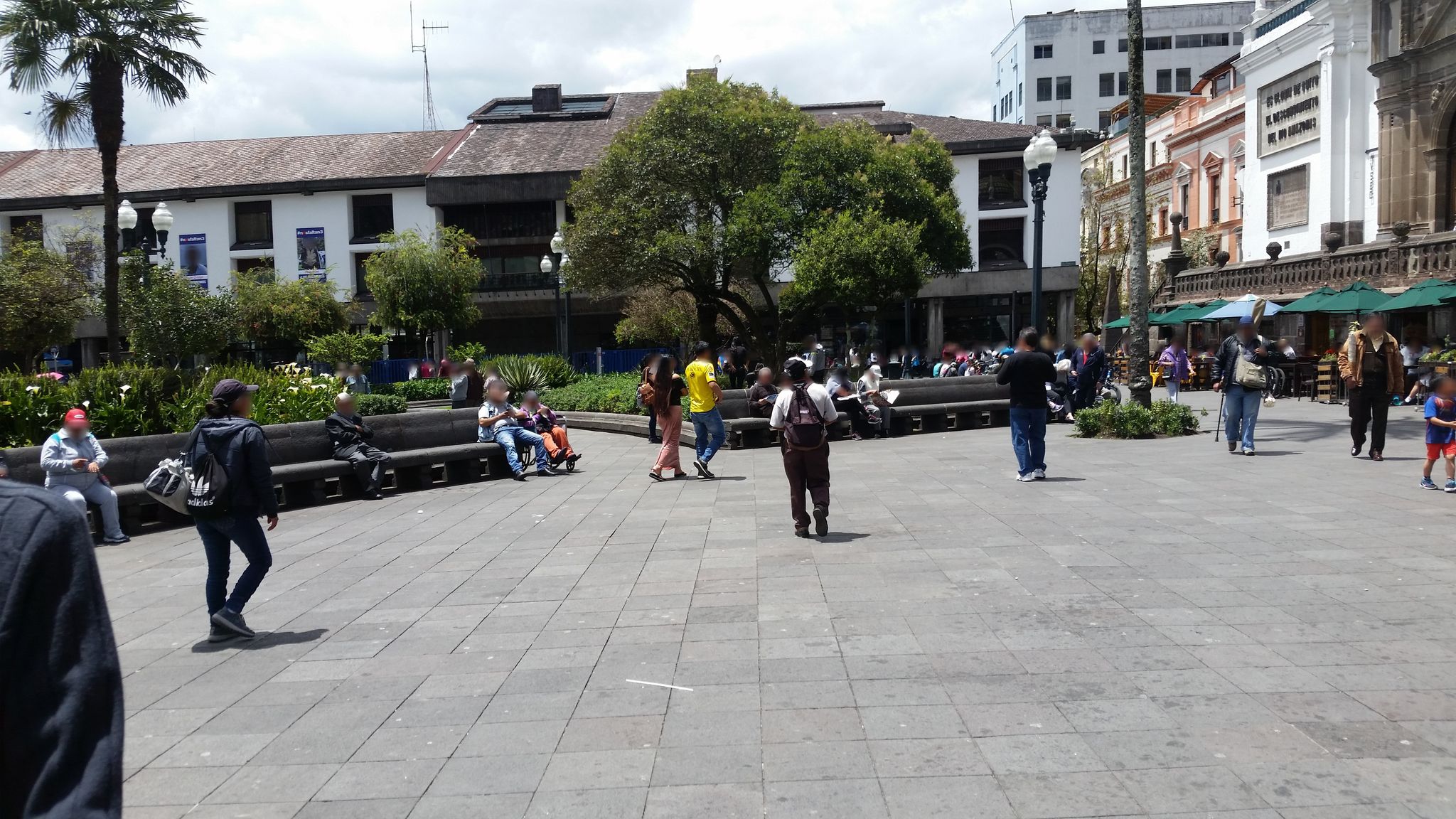}
    \vspace{-0.1cm} \caption[SVI from Quito, EC]{SVI from Quito, EC\footnoteref{mapillary} capturing sociability types.}
    \label{fig:quito-plaza} \vspace{-0.4cm}
\end{figure}

We illustrate the value of this taxonomy by analyzing an image from Quito, EC (Figure \ref{fig:quito-plaza}). In the scene, many individuals are seated on benches or at outdoor tables, passively observing the social environment---an example of passive sociability. Some of these seated individuals are interacting with passersby, likely acquaintances rather than close connections, representing fleeting sociability. Meanwhile, seated or walking couples exhibit enduring sociability, reflecting strong social ties. Categorizing social interactions in this way enables us to evaluate public spaces based on their intended social functions. For instance, this plaza may be designed to support passive sociability for people relaxing during a lunch break, whereas a restaurant might aim to foster enduring interactions, and a community center or outdoor sports facility could prioritize creating fleeting interactions and new social ties among neighbors. Without this taxonomy, we might simply count the number of people in each location and mistakenly conclude that they are equally socially active, overlooking the qualitative differences in social engagement.

\section{Inferring Social Context from SVI}

\subsection{Data Sources and Selection}

To explore whether SVI encodes meaningful social information, we used the National University Singapore (NUS) Global Streetscapes dataset \cite{HouGlobal2024}, which annotates Mapillary and KartaView images with metadata such as location, street type, time of day, season, and environmental context. We selected cities that overlap with municipalities from Round 7 of the World Values Survey (WVS) \cite{HaerpferWorld2022}, which provides independent, demographically representative measures of hundreds of qualitative variables across 64 countries administered every five years. We then took the average of the WVS responses for place attachment in each city. Images were filtered to match the WVS year for each city and to include only streets classified as ``walk,'' ``walk/cycle,'' or ``cycle'' in OpenStreetMap \cite{OpenStreetMap2017}, ensuring analysis of pedestrian-oriented spaces.  

After filtering, 15 cities remained: Ebina, JP, Bristol, EN, Trier, DE, Kasugai, JP, Manila, PH, Tokyo, JP, Dushanbe, TJ, Quito, EC, Sapporo, JP, Niigata, JP, Manchester, EN, Stuttgart, DE, Shizuoka, JP, Dresden, DE, and Hamamatsu, JP. From these, we randomly sampled 200 SVI to pair with the value for place attachment in each city. After dropping images with errors, our final dataset had 2,998 rows, with each row representing the annotations in each image. Values for place attachment in each city were applied to every image from that city. This strict filtering process yielded a workable sample size, underscoring the richness of SVI for studying public life. Future work can scale this approach to thousands of cities by extending beyond the pre-annotated NUS Global Streetscapes dataset.

\subsection{Classifying Social Interactions}

Our goal was not to develop a new computer vision model but to demonstrate that SVI can encode sociologically meaningful interaction patterns when interpreted through established social science theory. To our knowledge, this is the first study to translate a formal taxonomy of sociability into an automated street-view analysis.

Because no labeled dataset exists for social interaction in SVI, we used an MLLM as a practical tool to extract sociological information at scale. Directly using an MLLM is not a definitive method to extract this information; rather, it served as a proof-of-concept to unlock the latent social context already present in these images.

We use Mehta's taxonomy of social interactions \cite{MehtaStreets2019} to design a structured prompt\footnote{\url{https://anonymous.4open.science/r/SVI-prompt-2726/sociability_prompt_context.txt}} for ChatGPT-4o-mini \cite{OpenAIGPT4o2025}  to classify interactions in each image into passive, fleeting, and enduring sociability.

Before adopting the MLLM approach, we tested existing vision models, including a state-of-the-art social relation recognition model (GMLLM-SRR) \cite{TangCrossattention2025} trained on People in Social Context (PISC) \cite{LiDualglance2017}. However, these models failed to detect meaningful interactions due to the substantial distribution shift: PISC consists of close-up ``photo album'' images with clear facial expressions, whereas SVI features blurred faces, distant viewpoints, and repetitive activity contexts. Similarly, the only publicly available street view activity dataset \cite{HosseiniELSA2024} focuses on broad activity classes and does not distinguish between sociologically distinct interaction types.

By contrast, the MLLM-based approach successfully produced sociologically interpretable interaction labels across all sampled cities, demonstrating that even simple, off-the-shelf tools can operationalize established social science taxonomies at scale. As a representative example of MLLM outputs, Figure \ref{fig:quito-plaza} was classified as having ten passive sociability interactions, five fleeting sociability interactions, and two enduring sociability interactions. The next section relates these inferred interaction patterns to independent survey-based measures of place attachment.

\section{Relating SVI Interactions to Social and Built Environment Factors}

To assess whether the social context inferred from SVI reflects meaningful, real-world patterns, we tested correlations between the three inferred sociability types and established environmental and survey-based measures. Specifically, we used the WVS city-level place attachment score as an external indicator of community-level social cohesion and the Global Streetscapes green, sky, and water view indices as built-environment predictors. Linear regression models controlled for potential confounders, including the number of people in each image, time of day, and weather.

\subsection{Statistical Analysis: Proof of Concept}

After counting the number of occurrences of the three types of sociability for each image, we linked the results with WVS data for the corresponding city. Our hypothesis was that place attachment and built-environment features such as greenery and sky openness would predict higher levels of social interaction on the street.

Our dependent variables were the number of passive, fleeting, and enduring sociability interactions identified in each image, while the independent variables were the WVS city-level average for place attachment and the green, sky, and water view indices derived from the image segmentation pixel counts in the Global Streetscapes dataset \cite{HouGlobal2024}. These indices were calculated as the proportion of pixels representing each natural feature (i.e., green, sky, or water) relative to the number of pixels in the image. We controlled for potential confounders, including the number of people in the image, the weather in the image, and the time of day. 

For each of the three types of sociability, we created a separate linear regression model with the type of sociability as the dependent variable and the confounding variables, the view indices, and place attachment as independent variables. The statistical significance reported for each correlation is as reported by a linear regression from the \texttt{lm} function in R \cite{RCoreTeamLanguage2024}. Table \ref{tab:sociability-key} reports the model parameters and statistical significance for each sociability type. Notably, different types of social interaction had different occurrence rates: passive sociability had 3,744 examples across the dataset of 2,998 images, with fleeting sociability having 2,644 and enduring sociability having 942. 

\setlength{\tabcolsep}{2pt} 
\begin{table}[!b]
  \caption{Regression results for different types of sociability.}
  \label{tab:sociability-key}
  \centering
  \begin{tabular}{l@{\hskip 10pt}c@{\hskip 10pt}c@{\hskip 10pt}c}
    \toprule
    & Passive & Fleeting & Enduring \\
    \midrule
    Intercept      & -0.203         & 0.229          & -0.120         \\
                     & (0.289)        & (0.191)        & (0.091)        \\
    \makecell[l]{Green View Index} & 0.493\textsuperscript{+} & 0.106 & 0.221\textsuperscript{*} \\
                     & (0.287)        & (0.189)        & (0.090)        \\
    \makecell[l]{Sky View Index} & 1.088\textsuperscript{***} & 0.742\textsuperscript{***} & 0.432\textsuperscript{***} \\
                     & (0.284)     & (0.187)     & (0.090)     \\
    \makecell[l]{Water View Index} & 1.054 & 0.382 & 0.209 \\
                     & (3.124)        & (2.061)        & (0.985)        \\
    Place attachment             & 0.106         & 0.238\textsuperscript{*} & 0.078         \\
                     & (0.157)        & (0.103)        & (0.049)        \\
    \midrule
    Number of observations        & 2998           & 2998           & 2998           \\
    $R^2$            & 0.374          & 0.426          & 0.403          \\
    Adj. $R^2$       & 0.372          & 0.424          & 0.401          \\
    AIC              & 13002.5        & 10509.4        & 6084.4         \\
    BIC              & 13074.6        & 10581.5        & 6156.5         \\
    Log Likelihood   & -6489.274      & -5242.705      & -3030.209      \\
    RMSE             & 2.11           & 1.39           & 0.66           \\
    \bottomrule
  \end{tabular}

  \vspace{0.5em}
  {\begin{flushleft}
   \textit{Notes:}    \textsuperscript{***}$p < 0.001$; \textsuperscript{**}$p < 0.01$; \textsuperscript{*}$p < 0.05$; \textsuperscript{+}$p < 0.1$;  ($\cdots$) denotes standard error.
  \end{flushleft}
  }
\end{table}

\subsection{Interpretation of Results}

Several results align with long-standing urban sociology and environmental psychology theories, providing preliminary external validation for this approach:

\begin{itemize}
    \item Green view index was positively associated with enduring and passive social interactions. This is consistent with prior research showing that access to greenery promotes informal social interaction and casual co-presence, although this relationship has not been examined at the level of individual street images \cite{SullivanFruit2004}.

    \item Sky view index was correlated with all three sociability types. Whyte and Gehl have argued that built-environment features blocking views of the sky can create uncomfortable microclimates, discouraging people from lingering outdoors \cite{WhyteSocial1980, GehlCities1936}; our findings provide preliminary quantitative support for this theory across diverse cities.

    \item The view of water did not have any statistically significant relationship with social interaction. Prior research supports the idea that water features are mediators for social interaction in urban spaces \cite{WhyteSocial1980}, but we were not able to validate these findings.

    \item Place attachment was positively associated with fleeting sociability, consistent with theories that casual, everyday encounters contribute to stronger emotional bonds with one's city or neighborhood \cite{UjangLinking2018, LewickaWhat2010}. This external measure correlating with ChatGPT's classifications of sociability provides evidence that there is non-random data being extracted.

\end{itemize}

The results demonstrate that SVI encode latent social information that aligns with established social science theories. Even with a small, filtered dataset and a simple MLLM, we were able to detect theoretically consistent relationships between built-environment features, place attachment, and different types of sociability.

This should be interpreted as a proof of concept rather than definitive causal evidence. The goal was not to exhaustively test built-environment hypotheses, but to show that previously untapped, scalable visual data sources---when interpreted through social science theory---can yield meaningful sociological findings. That these simple, MLLM-based classifications align with established theories suggests that SVI can serve as a foundation for large-scale, cross-cultural research on urban sociability.

At the same time, the patterns we observed are conceptually intriguing because they imply that the built environment could be designed to facilitate specific types of interactions. Designing spaces for passive interactions might reduce social friction and ``othering'' in cities, designing for fleeting interactions could help connect communities across demographic lines, and designing for enduring interactions would strengthen bonds among families and close friends. Future studies—using larger datasets and more precise measures—could test these possibilities explicitly, advancing our understanding of how specific built-environment features shape different forms of social interaction.

\section{Future Work} 
SVI provides a unique opportunity for low-cost, privacy-preserving, distributed sensing of human behavior in built environments. The development of social relation recognition or emotional recognition datasets based on SVI data could support an unprecedented understanding of usage and experience in the urban environment. To provide some salient examples, urban planners could compare the vitality of cities and neighborhoods by the true social experience on the ground rather than the number of bodies in an area. Architects could quantify the social experiences their designs create. Environmental psychologists could take advantage of natural experiments occurring around the world and compare changes in social behavior as the built environment evolves. Public health researchers could directly calibrate agent-based pandemic models based on public behavior patterns in each municipality. Sociologists could visually see and estimate the strength of homophily effects by city and demographic. Unlocking the social potential of SVI will open a new chapter in the design of social, healthy, and livable cities.
\bibliographystyle{abbrvnat}
\bibliography{references}
\end{document}